\documentclass[sigconf]{acmart}

\usepackage{amsthm}
\usepackage{algorithm}
\usepackage{algorithmic}
\newlength\myindent
\usepackage[most]{tcolorbox}
\usepackage{multicol}

\AtBeginDocument{%
  \providecommand\BibTeX{{%
    \normalfont B\kern-0.5em{\scshape i\kern-0.25em b}\kern-0.8em\TeX}}}

\copyrightyear{2025}
\acmYear{2025}
\setcopyright{acmlicensed}\acmConference[GECCO '25 Companion]{Genetic and Evolutionary Computation Conference}{July 14--18, 2025}{Malaga, Spain}
\acmBooktitle{Genetic and Evolutionary Computation Conference (GECCO '25 Companion), July 14--18, 2025, Malaga, Spain}
\acmDOI{10.1145/3712255.3736397}
\acmISBN{979-8-4007-1464-1/2025/07}

\begin{document}
\pagestyle{plain}

\title{GPU-Accelerated Rule Evaluation and Evolution}
\author{Hormoz Shahrzad}
\email{hshahrzad@cs.utexas.edu}
\orcid{0000-0002-5983-4531}
\affiliation{%
  \institution{The University of Texas at Austin;}
%  \city{Austin}
 \country{USA}
}
\affiliation{%
  \institution{Cognizant AI Labs,}
  \city{San Francisco}
  \state{California}
  \country{USA}
}
\author{Risto Miikkulainen}
\email{risto@cs.utexas.edu}
\orcid{0000-0002-0062-0037}
\affiliation{%
  \institution{The University of Texas at Austin;}
%  \city{Austin}
 \country{USA}
}
\affiliation{%
  \institution{Cognizant AI Labs,}
  \city{San Francisco}
  \state{California}
  \country{USA}
}

\renewcommand{\shortauthors}{Shahrzad and Miikkulainen}


\begin{abstract}
This paper presents an extension to the EVOTER (Evolution of Transparent Explainable Rule Sets) framework, previously introduced to promote transparency and explainability in AI through the evolution of interpretable rule sets. While EVOTER demonstrated strong performance in generating compact and understandable rule-based models using a list-based grammar, the current work, Accelerated Evolutionary Rule-based Learning (AERL), builds upon this foundation by addressing scalability challenges. Specifically, AERL leverages GPU-accelerated computation and back-propagation fine-tuning within a PyTorch framework to significantly enhance the efficiency of rule evaluation and evolution. By introducing a tensorized numerical representation of rules and incorporating gradient-based optimization, AERL achieves superior computational speed and improved search-space exploration. Experimental results confirm that the approach is effective, achieving faster convergence and similar accuracy. It thus offers a scalable path forward for explainable AI.
\end{abstract}

%\begin{CCSXML}
%<ccs2012>
%<concept>
%<concept_id>10010147.10010257.10010293.10011809.10011812</concept_id>
%<concept_desc>Computing methodologies~Genetic algorithms</concept_desc>
%<concept_significance>500</concept_significance>
%</concept>
%<concept>
%<concept_id>10010147.10010257.10010293.10011809.10011813</concept_id>
%<concept_desc>Computing methodologies~Genetic programming</concept_desc>
%<concept_significance>300</concept_significance>
%</concept>
%</ccs2012>
%\end{CCSXML}
%
%\ccsdesc[500]{Computing methodologies~Genetic algorithms}
%\ccsdesc[300]{Computing methodologies~Genetic programming}

%\keywords{Genetic Algorithms, rule set Evolution, Explainable AI, XAI}


\maketitle

\vspace{-1ex}
\section{Introduction}
\label{sc:introduction}
Evolutionary Rule-based Machine Learning (ERL) systems are extensively used in supervised learning, where rule fitness is typically measured by classification accuracy on labeled datasets~\cite{miikkulainen:emlchapter23}. A recent contribution in this area is EVOTER, which evolves expressive rule sets using a list-based grammar~\cite{EVOTER-2204-10438}.

This work extends EVOTER by addressing a key bottleneck in ERL: the high cost of rule evaluation. GPUs have long accelerated optimization methods—including Mixed Integer Non-Linear Programming (MINLP), Central Force Optimization (CFO), Genetic Algorithms (GA), and Particle Swarm Optimization (PSO)—achieving speedups from 7$\times$ to 10,000$\times$ ~\cite{Singh2014ACS}, with promising results reported across domains~\cite{Jhne2016OverviewOT}. Building on this potential, this paper proposes a tensor-based rule representation in PyTorch, enabling parallel GPU evaluation and gradient-based fine-tuning via back-propagation for rule-based systems.

The proposed approach, Accelerated Evolutionary Rule-based Learning (AERL), improves scalability and enhances search efficiency. Experiments on the UCI Breast Cancer dataset demonstrate significant speed gains.

\begin{figure}[!t]
\footnotesize
\centering

\begin{tcolorbox}[colframe=black!60, colback=white, boxrule=0.4pt, sharp corners,
boxsep=0pt, left=1pt, right=1pt, top=1pt, bottom=1pt, width=0.95\linewidth]
\textbf{Features} (only six included for simplicity): \\
\textit{mean radius, fractal dimension error, worst compactness,} \\
\textit{worst symmetry, perimeter error, mean area} \\
\textbf{Rule 1:} (Rule 1 has three conditions) \\
Conditions: \text{tensor([[1.0, 0, 0, 0, 0, 0], [0, 1.0, -0.38, 0, 0, 0], [0, 0, 0, 1.0, 0, 0]])}, \\
Biases: \text{tensor([-2.02, 0.0, -0.29])}, \# 0.0 shows feature-to-feature comparison \\
\textbf{Rule 2:} (Rule 2 has two conditions)\\
Conditions: \text{tensor([[0, 0, 0, 0, 1.0, 0], [0, 0, 0, 0, 0, 1.0]])} \\
Biases: \text{tensor([-0.43, -0.29])}
\end{tcolorbox}
\vspace{-0.5ex}
{\footnotesize ($a$) Numerical representation of the rule set before fine-tuning\par}

\vspace{0.6ex}
\begin{tcolorbox}[colframe=black!60, colback=white, boxrule=0.4pt, sharp corners,
boxsep=0pt, left=1pt, right=1pt, top=1pt, bottom=1pt, width=0.95\linewidth]
\begin{flushleft}
1. (\textit{mean radius} $>$ 2.02) \& \\
\phantom{1.}(\textit{fractal dimension error} $>$ 0.38 $\times$ \textit{worst compactness}) \& \\
\phantom{1.}(\textit{worst symmetry} $>$ 0.29) $\longrightarrow$ \textit{malignant}, \\
2. (\textit{perimeter error} $>$ 0.43) \& \\
\phantom{2.}(\textit{mean area} $>$ 0.29) $\longrightarrow$ \textit{benign}
\end{flushleft}
\end{tcolorbox}
\vspace{-0.5ex}
{\footnotesize ($b$) Rule set before fine-tuning\par}

\vspace{0.6ex}
\begin{tcolorbox}[colframe=black!60, colback=white, boxrule=0.4pt, sharp corners,
boxsep=0pt, left=1pt, right=1pt, top=1pt, bottom=1pt, width=0.95\linewidth]
\begin{flushleft}
1. (\textit{mean radius} $>$ 0.06) \& \\
\phantom{1.}($-$\textit{fractal dimension error} $>$ $-$7275.88 $\times$ \textit{worst compactness}) \& \\
\phantom{1.}(\textit{worst symmetry} $>$ $-$3.69) $\longrightarrow$ \textit{malignant}, \\
2. ($-$\textit{perimeter error} $>$ $-$0.98) \& \\
\phantom{2.}($-$\textit{mean area} $>$ $-$0.24) $\longrightarrow$ \textit{benign}
\end{flushleft}
\end{tcolorbox}
\vspace{-0.5ex}
{\footnotesize ($c$) The same rule set after fine-tuning\par}

\captionsetup{font=scriptsize}
\vspace{-2ex}
\caption{\emph{Example of rule set representation and fine-tuning.} A randomly generated rule set (not evolved) is shown in both numerical (tensor) form and human-readable form to illustrate how feature weights and biases translate into logical conditions. Back-propagation is applied to tune the coefficients, demonstrating how fine-tuning alone can substantially improve accuracy while keeping the logical structure intact.}
\label{fg:ruleset}
\vspace{-4ex}
\end{figure}

\vspace{-1ex}
\section{Background}
\label{sc:background}

Evolving transparent, explainable rule sets using extended propositional logic is a promising approach in explainable AI (XAI), offering interpretability that black-box models lack~\cite{EVOTER-2204-10438, bp-sepsis}. While effective for compact rule discovery, scalability remains a challenge, particularly during computationally intensive evaluation phases.

GPUs have been widely used to accelerate evolutionary machine learning (EML). One such area is Learning Classifier Systems (LCS \cite{urbanowicz2009learning} and especially XCS \cite{kovacs2001what}). When individual rules are evaluated and evolved within a population through the Michigan-style approach, GPU acceleration has shown promise in this process~\cite{butz2011fast, tan2013gpu}. More generally, techniques have been developed for speeding up evaluations in general evolutionary learning on general-purpose GPUs~\cite{franco2010speeding}. Recent surveys and meta-analyses~\cite{heider2023metaheuristic, siddique2024survey} emphasize growing interest in efficient hybrid symbolic-subsymbolic systems as well.

Similar gains have been more difficult to achieve in the Pittsburgh-style approach,~\cite{cano2014parallel}, where each individual represents a complete rule set and is evaluated holistically. The structural mismatch between symbolic rules and GPU-optimized numerical formats is more severe, making it difficult to parallelize the evaluations.  This paper introduces a tensorized rule representation in PyTorch, enabling efficient GPU-based evaluation and backpropagation-based fine-tuning, and offering a Pittsburgh-style alternative to population-level approaches.

\vspace*{-1ex}
\section{Method}
\label{sc:method}

AERL transforms list-based propositional logic expressions~\cite{EVOTER-2204-10438} into a tensorized form amenable to linear algebra and optimization. Rules are represented as inequalities, allowing matrix multiplication to evaluate multiple rules in parallel (Figure~\ref{fg:ruleset}$a$).
Each row in a rule’s tensor corresponds to a condition, with non-zero entries indicating feature weights and the associated bias representing the threshold. For example, a row\\ \centerline{\texttt{[0, 1.0, -0.38, 0, 0, 0]}}
with a bias of 0 encodes the inequality\\
\centerline{\textit{fractal dimension error} > 0.38 × \textit{worst compactness}.}
This mapping is shown in Figure~\ref{fg:ruleset}$b$, which translates the tensor form into a human-readable rule set.
For inference, an activation function clips negative values to zero (false) and rounds positives to one (true), enabling binary decisions from continuous outputs.

The second innovation in AERL is fine-tuning with backpropagation. Since evaluation already occurs on GPUs, it is a way to take advantage of them more fully. A differentiable activation function (e.g., sigmoid) provides gradients for back-propagation, making it possible to adjust rule coefficients without altering logical structure.

Evolution proceeds via tournament selection and offspring generation with randomized coefficients. Back-propagation refines fitness at each generation, guiding evolution through the Baldwin effect~\cite{baldwin-1896}. Evolution runs on CPUs, with GPU experiments loading candidates for accelerated evaluation.

Figures~\ref{fg:ruleset}$b$ and~\ref{fg:ruleset}$c$ illustrate a random example rule set (not evolved) for the UCI Breast Cancer dataset: its accuracy improved from 21.05\% to 92.40\% after 50 epochs of fine-tuning. This example highlights how coefficient optimization alone—without evolutionary search—can substantially boost accuracy while keeping the logical structure intact.

\vspace{-1ex}
\section{Experimental Results}
\label{sc:experiments}
%An over five-fold speedup was achieved with GPUs, and a statistically significant improvement in accuracy with fine tuning.

\subsection{Experiment Setup}
\label{sc:experiments-setup}
The experiments aim to demonstrate computational efficiency gains from the tensorized rule representation and GPU acceleration, using the UCI Breast Cancer dataset as a well-known and accessible benchmark~\cite{misc_breast_cancer_14}. As the focus is on acceleration rather than classification accuracy, this dataset suffices to illustrate the method’s effect; the approach is compatible with EVOTER and systems using similar grammars, and can be readily scaled to larger datasets.

Experiments compared GPU vs.\ CPU performance with and without back-propagation-based fine-tuning. Each setup was run 10 times with a fixed pool of 100 candidate rule sets on Google Colab T4-GPU. Fine-tuning applied 50 epochs of gradient updates to rule coefficients.

\begin{figure}[!t]
\begin{minipage}[b]{\linewidth}
\centering
\includegraphics[width=0.75\textwidth]{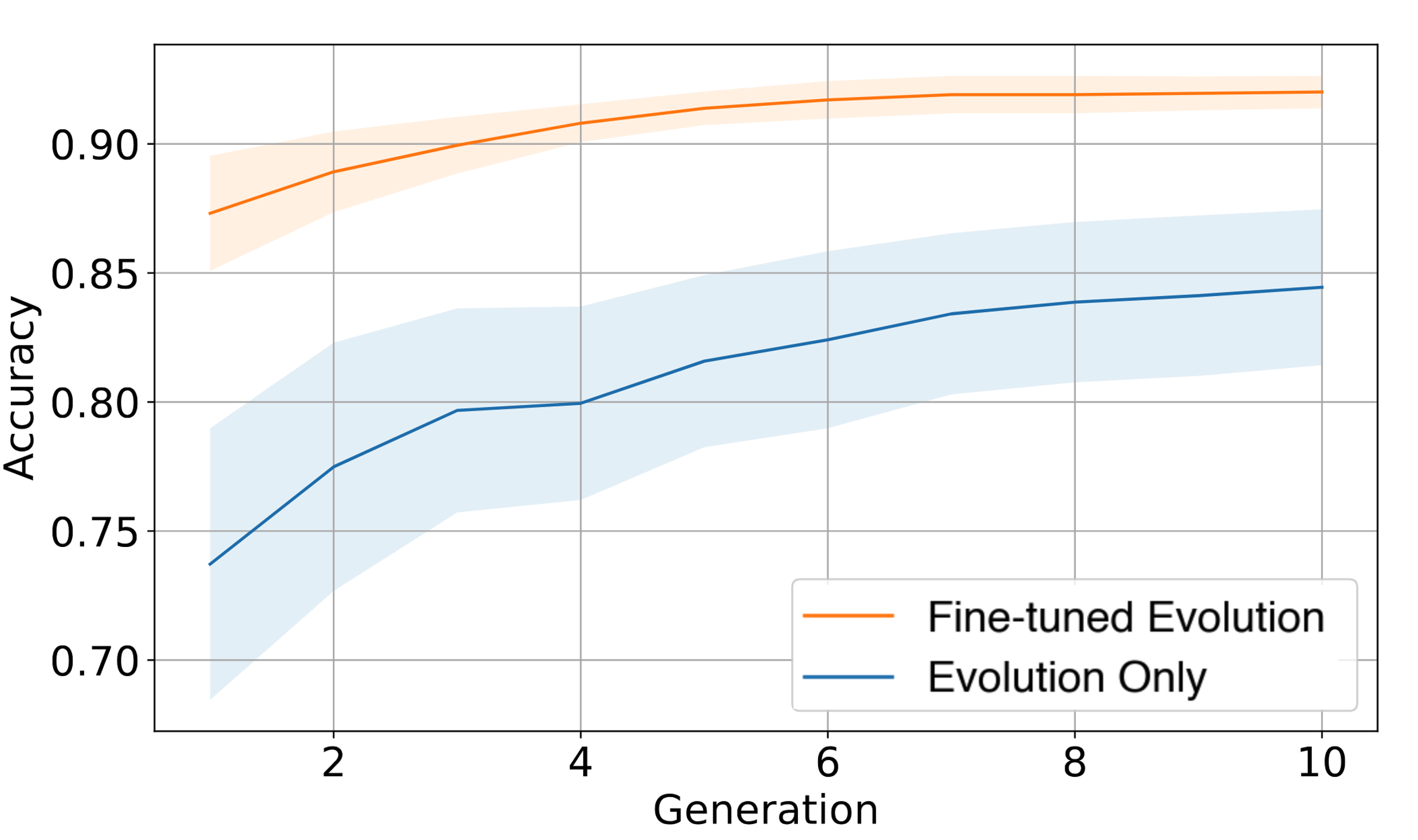}
\vspace*{-2ex}
    \captionsetup{font=scriptsize}
    \caption{\emph{Evolutionary Progress of PyTorch Rules in the UCI Breast Cancer Domain with vs. without Fine-tuning.} Each point is averaged over 10 runs, with 95\% confidence intervals. In both cases, rule structures are evolved, but in the fine-tuning setting, back-propagation is applied at each generation to optimize rule coefficients without altering structure. This gradient-based refinement improves accuracy and accelerates search via the Baldwin effect ($p = 2 \times 10^{-7}$).}
    \label{fig:accuracy_comparison}
    \end{minipage}
\vspace*{-6ex}
\end{figure}

\begin{figure}[!t]
\begin{minipage}[b]{\linewidth}
\centering
\includegraphics[width=0.75\textwidth]{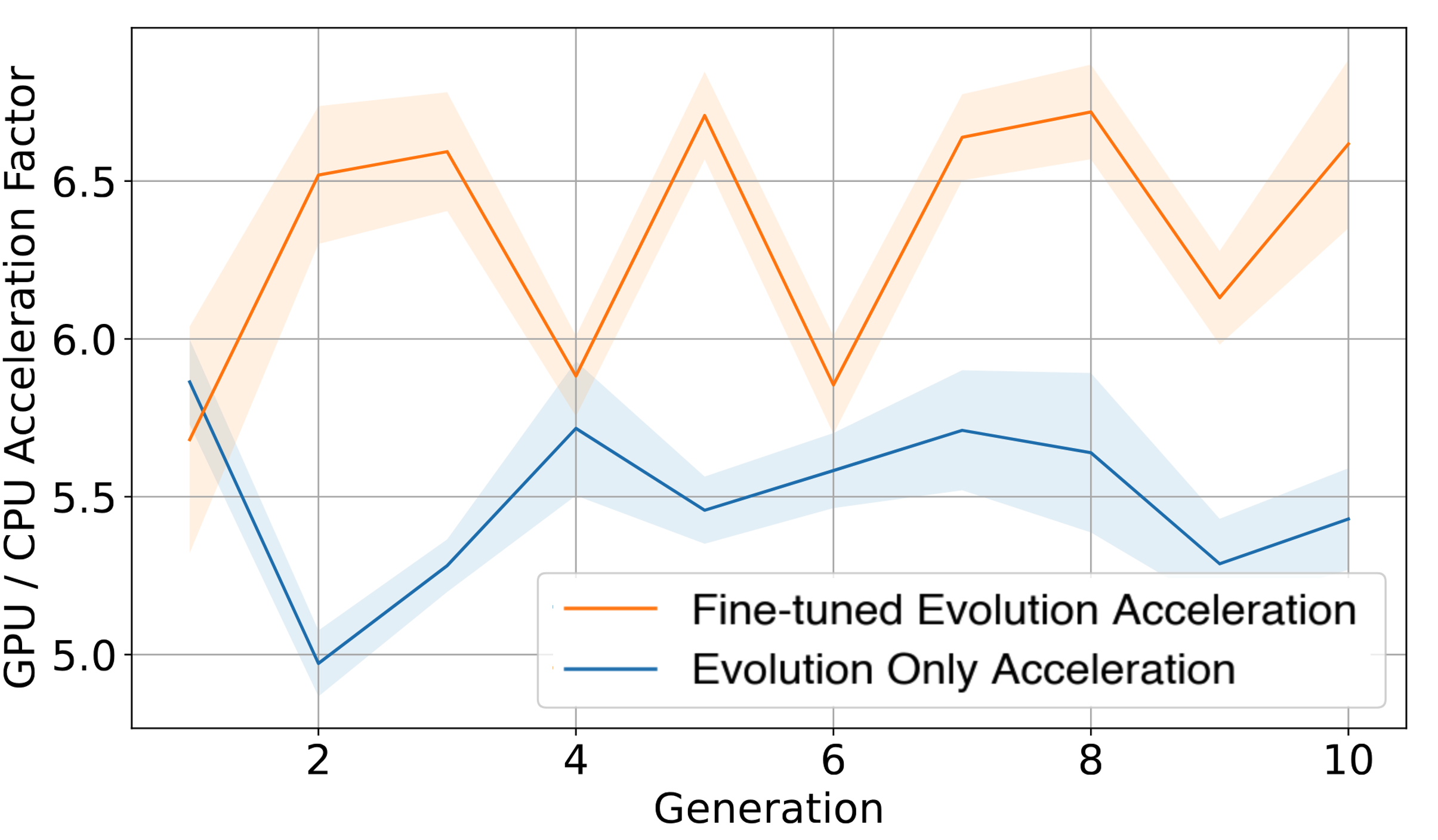}
\vspace*{-2ex}
    \captionsetup{font=scriptsize}
    \caption{\emph{Comparison of GPU to CPU Acceleration Ratios with and without Fine-tuning.} Plots show average CPU/GPU wallclock time ratio over 10 runs with 95\% confidence intervals. Both with and without fine-tuning, GPU acceleration provides a minimum fivefold speedup compared to CPUs. Observed variations are due to the total number of rules and conditions per generation within a fixed pool size of 100 candidates. These results demonstrate that GPUs significantly accelerate rule-set evolution with and without fine-tuning.}
    \label{fig:acceleration_comparison}
    \end{minipage}
\vspace*{-5ex}
\end{figure}

\subsection{Evolutionary Performance}
\label{sc:experiments-performance}
The performance of the PyTorch rules was first measured in terms of accuracy improvements over generations. As shown in Figure~\ref{fig:accuracy_comparison}, the application of fine-tuning significantly expedited the evolutionary progress, achieving statistically significant performance improvements ($p=2 \times 10^{-7}$). This result highlights how the method leaves the search for logical structure to evolution while reducing the search space by fine-tuning rule coefficients to their best values within each structure.

\subsection{Computational Efficiency}
Figure \ref{fig:acceleration_comparison} illustrates the acceleration ratios achieved through GPU computation. Notably, both with and without fine-tuning, the use of GPU resources consistently resulted in at least a fivefold speed increase over traditional CPU processing. This acceleration was observed despite variations in the number of rules and conditions per generation, highlighting the efficacy of GPU acceleration in expediting the evolutionary computation of PyTorch rules.

\vspace{-1ex}
\section{Discussion and Future Work}
\label{sc:discussion}

The proposed AERL framework offers a scalable approach to rule-set evolution by leveraging GPU acceleration without compromising model quality—results from CPU and GPU implementations are equivalent in terms of classification accuracy, with performance gains stemming solely from faster computation.

Future improvements include extending support for categorical features to broaden applicability across diverse domains. Incorporating advanced rule structures, such as time-lags and exponents, may capture richer temporal or nonlinear dependencies. Probabilistic outputs can be enabled via Softmax activation, offering finer-grained class predictions~\cite{Hodjat2018}. 

\newpage
\bibliographystyle{ACM-Reference-Format}
\bibliography{main}

\end{document}